\documentclass[letterpaper, 10 pt, conference]{ieeeconf}  
\IEEEoverridecommandlockouts                              

\usepackage{subcaption}
\usepackage{graphicx}

\usepackage{xcolor}

\definecolor{darkteal}{RGB}{0, 125, 121}
\definecolor{lightteal}{RGB}{0, 93, 93}

\usepackage{mathptmx} 
\usepackage{times} 
\usepackage{amsmath} 
\usepackage{amssymb}  
\usepackage{mathtools}

\usepackage{booktabs}

\usepackage{hyperref}

\usepackage[font={small}]{caption}

\usepackage{microtype}

\usepackage{float}

\title{\LARGE \bf
PainDiffusion: Learning to Express Pain
}

\author{Quang Tien Dam$^{1}$, Tri Tung Nguyen Nguyen$^{1}$, Yuuki Endo$^{1}$, Dinh Tuan Tran$^{3}$ and Joo-Ho Lee$^{2}$
\thanks{$^{1}$Graduate School of Information Science and Engineering, Ritsumeikan University, Japan. {\tt\small tien.aislab@gmail.com}
        }%
\thanks{$^{2}$College of Information Science and Engineering, Ritsumeikan University, Japan.
        {\tt\small leejooho@is.ritsumei.ac.jp}}%
\thanks{$^{3}$Faculty of Data Science, Shiga University, Japan.}
}

\begin{document}

\maketitle



\begin{abstract}

Accurate pain expression synthesis is essential for improving clinical training and human-robot interaction. Current Robotic Patient Simulators (RPSs) lack realistic pain facial expressions, limiting their effectiveness in medical training. In this work, we introduce PainDiffusion, a generative model that synthesizes naturalistic facial pain expressions. Unlike traditional heuristic or autoregressive methods, PainDiffusion operates in a continuous latent space, ensuring smoother and more natural facial motion while supporting indefinite-length generation via diffusion forcing. Our approach incorporates intrinsic characteristics such as pain expressiveness and emotion, allowing for personalized and controllable pain expression synthesis. We train and evaluate our model using the BioVid HeatPain Database. Additionally, we integrate PainDiffusion into a robotic system to assess its applicability in real-time rehabilitation exercises. Qualitative studies with clinicians reveal that PainDiffusion produces realistic pain expressions, with a 31.2\% ± 4.8\% preference rate against ground-truth recordings. Our results suggest that PainDiffusion can serve as a viable alternative to real patients in clinical training and simulation, bridging the gap between synthetic and naturalistic pain expression. Code and videos are available at: \href{https://damtien444.github.io/paindf/}{https://damtien444.github.io/paindf/}.

\end{abstract}

\section{INTRODUCTION}


Reading patient pain accurately is crucial for improving clinical care \cite{henry_association_2012}. However, research indicates that clinicians often underestimate or misinterpret patient pain compared to laypeople, possibly due to cognitive biases or overreliance on medical instruments \cite{prkachin_expressing_1995, giannini_measurement_2000}. This underestimation can lead to inadequate pain management, misdiagnosis, and patient distress, increasing health risks \cite{jansen_emotional_2010}.

Pain is a multimodal phenomenon involving biological signals, facial expressions, heart rate, skin color changes, speech tone, and more \cite{mcguire_comprehensive_1992}. Among these, facial expressions provide critical nonverbal cues that help clinicians assess pain intensity and emotional distress. Training clinicians to recognize these expressions more effectively could improve patient care \cite{moosaei_using_2017}.

Robotic Patient Simulators (RPSs) allow healthcare professionals to practice procedures and diagnostic skills without risking patient harm. Current RPSs can simulate limb movements, breathing, bleeding, and biosignals, but they often lack realistic facial expressions \cite{pourebadi_facial_2022}. Given that 70\% of medical errors stem from communication issues, and 75\% of those lead to patient's death \cite{leonard_human_2004}, enhancing RPSs with realistic facial reactions could significantly improve that communication.

\begin{figure}[t]
    \centering
    \includegraphics[width=1\linewidth]{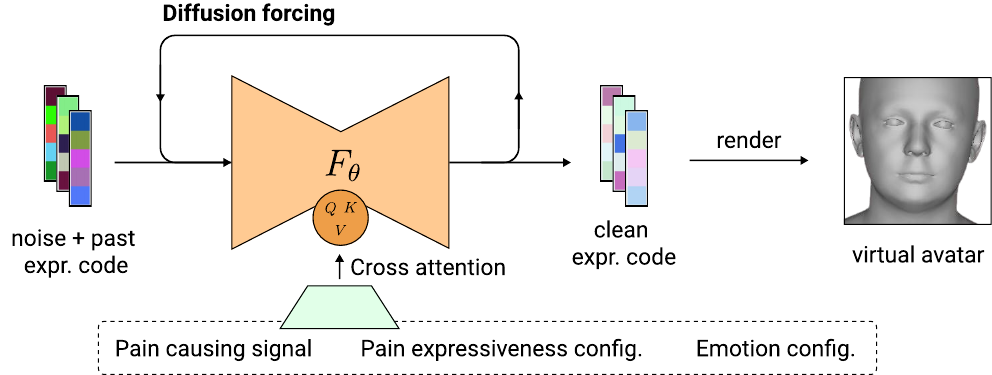}
    \small\caption{\textbf{Overview.} PainDiffusion inputs pain stimuli signals, expressiveness configuration, emotion status, and past frames to generate the next appropriate pain facial reaction.}
    \label{fig:model-flow}
\end{figure}

A major challenge in incorporating facial pain expressions into RPSs is the high-dimensional, nonlinear mapping between pain stimuli and facial responses. Pain expressions are inherently probabilistic and modulated by inter-individual factors such as demographic attributes, baseline expressivity, emotional state, and pain type. Traditional heuristic or rule-based systems fail to generalize across such complexity, as they rely on static, predefined mappings that do not capture the stochastic nature of facial responses \cite{moosaei_using_2017, moosaei_naturalistic_2014, lee2021care}.

We hypothesize that leveraging a naturalistic, non-acted pain dataset can better model the variability and uncertainty inherent in pain expressions. Recent advancements in deep generative models, including diffusion models, variational autoencoders (VAEs), offer a principled approach for learning the latent structure of facial pain expressions. By conditioning generative models on multimodal pain-related signals (e.g., physiological markers, stimulus properties, and affective states), we can develop a data-driven system that synthesizes realistic facial expressions in response to dynamic pain stimuli. This approach bypasses the limitations of manually designed animation heuristics, instead enabling adaptive, personalized facial expression generation that aligns with real-world pain perception dynamics.

In this work, we focus on synthesizing facial reactions based on embodiment signals. We exclusively use the BioVid HeatPain Database \cite{walter_biovid_2013}, which is the only dataset that includes direct recordings of stimuli, physiological signals, and naturalistic facial reactions. This restricts our ability to generalize across diverse patient populations and cultural backgrounds. Due to limited resources, we can currently only deploy the model on virtual avatars, reserving real-world deployment for future work. Our contributions are threefold: (1) We introduce PainDiffusion, a model designed to generate pain-related facial expressions with arbitrary-length predictions, making it suitable for robotic applications. It incorporates intrinsic characteristics such as expressiveness and emotion, allowing for more controllable and personalized generation. (2) We propose a new set of baselines and metrics to effectively evaluate the quality and accuracy of pain expressions generated by our model. (3) Finally, we integrate PainDiffusion with a robotic elbow and have rehabilitation clinicians assess the pain reactions to their actions.

\section{RELATED WORKS}

In training robots, earlier work in pain synthesizing primarily focused on recognizing pain situations and selecting expressions from a predefined set \cite{lee2021care, haque_deep_2018}. Among pain-related research, most efforts have focused on recognition and classification using either traditional machine learning or deep learning techniques \cite{ huang2019pain, moosaei_naturalistic_2014, moosaei_using_2017}. However, those methods often result in expressions that are \textit{unnatural and not automatic}. More recently, research has shifted towards using VQ-VAE and autoregressive models for facial expressions \cite{ng_learning_2022}; however, their decoders can produce jittery motion at the boundaries of the motion sequence. In human action generation, recent work is increasingly focusing on diffusion models to achieve state-of-the-art results \cite{chen_humanmac_2023, tian_transfusion_2024, barquero_belfusion_2023}, which generate motions in a \textit{seamless} way. However, these models typically assume fixed-length window generation and require many forward passes, making them unsuitable for robotic applications. Therefore, to address the lack of naturalness in synthesized pain expressions, this paper proposes a novel approach using diffusion models to generate realistic, seamless, and automatic pain behaviors for robots.

We ultimately chose diffusion models for four key reasons. First, they operate in a continuous data domain, enabling smoother and more natural facial motion—an area where autoregressive models struggle \cite{barquero_belfusion_2023, kirschstein_diffusionavatars_2024}. Second, diffusion models support diffusion forcing, allowing for indefinite-length signal generation without divergence, a challenge faced by both GAN-based and autoregressive approaches \cite{chen_diffusion_2024}. Third, they have demonstrated high-quality performance across multiple modalities as mentioned previously, making them a robust choice for pain expression synthesis. Finally, diffusion models offer controllability over the influence of different conditioning signals through classifier-free guidance, enhancing their adaptability to diverse use cases \cite{ho_classifier-free_2022}.

\section{PAIN DIFFUSION}

Our main goal is to model the relationship between facial expressions, pain-causing signals, and the intrinsic features of an individual. To achieve this, we define the high-level task as follows: \textit{Given a continuous sequence of pain-causing signals and the configuration of the individual, we autoregressively predict the appropriate facial expression.}

To capture ongoing reactions, we employ diffusion forcing (Sec. \ref{dforcing}) to roll a denoising diffusion model (Sec. \ref{ediff}) to generalize the prediction further than the training temporal horizon. We developed a temporal latent U-Net (Sec. \ref{tlunet}) with temporal attention to enhance temporal coherence in the predictions. This model can process a sequence of conditioning signals and intrinsic configurations, generating a latent vector representing the output produced by EMOCA \cite{danecek_emoca_2022}, as described in Sec. \ref{f-repre}. The high-level system is illustrated in Fig. \ref{fig:model-flow}.

\begin{figure}
    \centering
    \includegraphics[width=\linewidth]{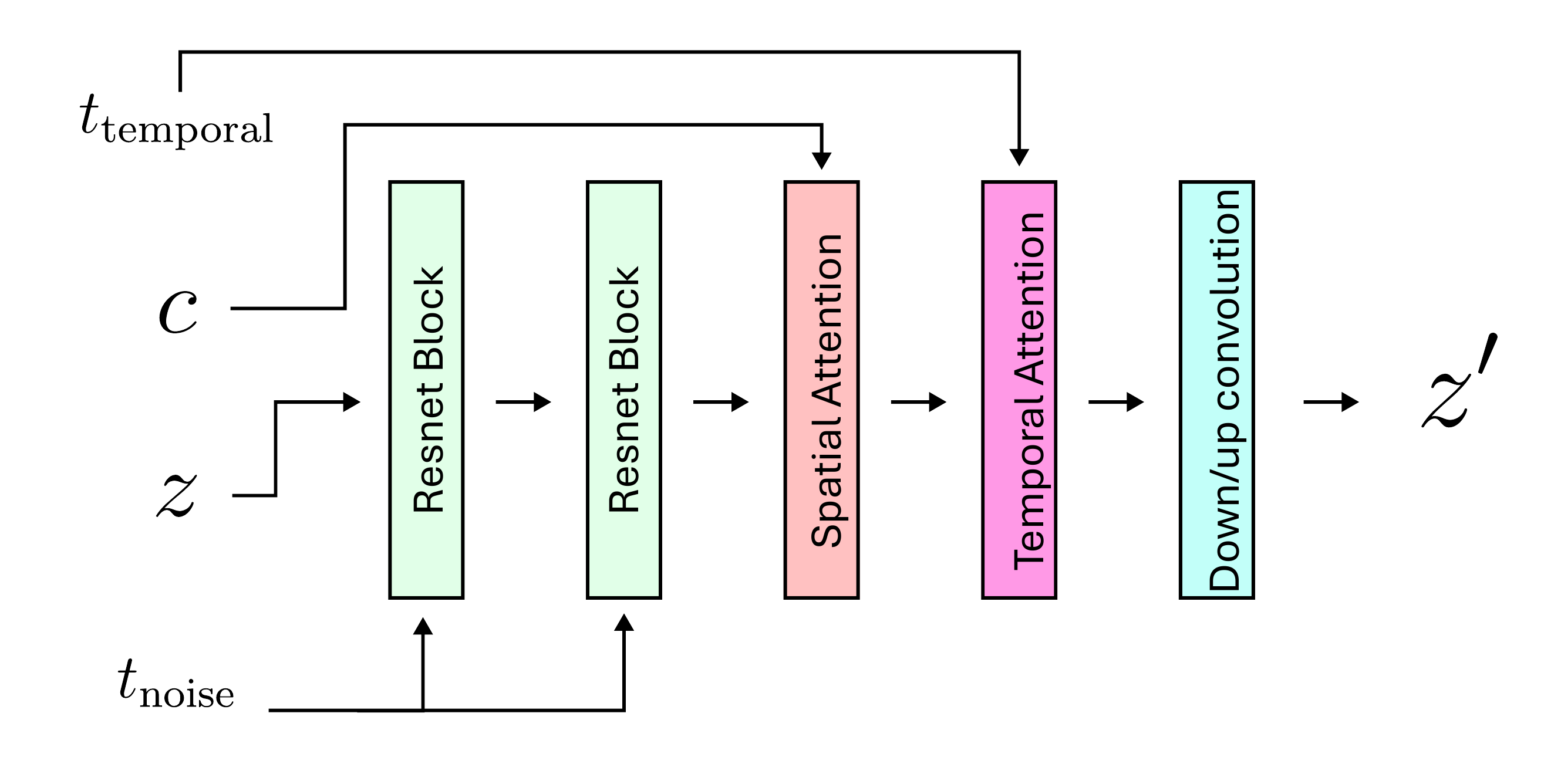}
    \caption{\textbf{U-Net Blocks.} The Temporal U-Net block incorporates \( t_{\text{noise}} \) into the Convolution 1D ResNetBlock using a scale-shift operation. Next, spatial attention applies cross-attention to integrate the condition information \( c \). Temporal time \( t_{\text{temporal}} \) is embedded using sinusoidal embeddings, followed by cross-attention in the temporal attention block to help the model understand temporal dynamics. Finally, the features are scaled up or down. The skip connection in the up blocks is concatenated with \( z \) from the down blocks.}
    \label{fig:unet-block}
\end{figure}

\subsection{Problem Definition}

Let \( y \in \mathbb{R}^d \) represent a facial expression, where \( Y = (y_0, y_1, \ldots, y_n) \) is a sequence of facial expressions with length \( n \). In our approach, the generation of \( Y \) is conditioned on a pain-causing signal \( C = (c_0, c_1, \ldots, c_n) \), where each \( c_i \in \mathbb{R} \). Additionally, the model is guided by a pain expressiveness configuration parameter \( \mathcal{P} \in \mathbb{R} \), which controls the intensity of the pain expression, allowing the model to capture individual differences in how pain is expressed. Since a person’s pain expression may vary depending on their current emotional state, even under identical pain stimuli, we introduce an emotion configuration parameter \( \mathcal{E} \in \mathbb{R} \). This parameter is included to adjust the generated facial expressions to account for the influence of the subject's emotional state during the pain expression sequence.

\subsection{Facial Representations}
\label{f-repre}

We utilize EMOCA \cite{danecek_emoca_2022} to produce 3D latent codes as it effectively maps from a disentangled latent space to high-quality face meshes with a reasonable render time. EMOCA \cite{danecek_emoca_2022} builds on the FLAME \cite{li_learning_2017} 3D face mesh model and DECA's \cite{feng2021learning} method of decomposing an image \( I \) into factors such as shape, albedo, and lighting, but places greater emphasis on maintaining emotion consistency in the output. The EMOCA latent representation is modeled as:

\begin{equation}
    E_c(I) \rightarrow (\boldsymbol{\beta}, \boldsymbol{\theta}, \boldsymbol{\psi}, \boldsymbol{\alpha}, \mathbf{l}, \mathbf{c}),
\end{equation}

where \( \boldsymbol{\beta} \in \mathbb{R}^{|\beta|} \) represents identity shape, \( \boldsymbol{\theta} \in \mathbb{R}^{|\theta|} \) are pose parameters, \( \boldsymbol{\psi} \in \mathbb{R}^{|\psi|} \) represents facial expressions, \( \boldsymbol{\alpha} \) is albedo, \( \mathbf{l} \in \mathbb{R}^{27} \) represents Spherical Harmonics lighting, and \( \mathbf{c} \in \mathbb{R}^3 \) represents camera parameters. In this approach, we focus specifically on facial expression and jaw pose, so the latent space is reduced to the concatenation \( (\boldsymbol{\psi}, \boldsymbol{\theta_{\text{jaw}}}) \), while other features are fixed as the average values from the training dataset. By excluding less relevant information such as lighting conditions and other constant features, the model can concentrate on capturing the dynamics of facial expression changes.

After generating the new facial expression and jaw pose \( (\boldsymbol{\psi}, \boldsymbol{\theta_{\text{jaw}}})_{\text{pred}} \) , the 2D face image \( I_{\text{pred}} \) can be rendered using the render function \( R \) from PyTorch3D \cite{pytorch3d} along with the FLAME model \( M \):

\begin{equation}
    \mathcal{R}(\boldsymbol{\psi}, \boldsymbol{\theta_{\text{jaw}}}) = R(M(\boldsymbol{\beta}, \boldsymbol{\theta}_{\text{pred}}, \boldsymbol{\psi}_{\text{pred}}), \boldsymbol{\alpha}, \mathbf{l}, \mathbf{c}) \rightarrow I_{\text{pred}}.
\end{equation}

This rendering process allows the model to synthesize the predicted facial expression as a 2D image by leveraging the 3D mesh generated by the FLAME model. During the image render state, we learn that EMOCA latent space has a small variance for the pose parameter, which makes the diffusion model unable to focus on generating stable poses across frames, we scale the feature bigger to have the same variance with other parameters, then scale it down to render with EMOCA.




\subsection{Temporal Latent Unet}
\label{tlunet}






Latent diffusion models (LDMs) have been effectively utilized for generating images, videos, and human behaviors \cite{blattmann_align_2023, barquero_belfusion_2023, kirschstein_diffusionavatars_2024, xu_vasa-1_2024}. In our work, we follow the LDM approach by restricting data representation to the EMOCA latent space, resulting in a much smaller data space compared to conventional diffusion models for image or video generation. Prior works in video generation diffusion models \cite{ho_video_2022, blattmann_align_2023} have introduced temporal layers to the standard spatial U-Net \cite{unet_2015} to better capture temporal information. Instead of employing 2D convolutions as in \cite{rombach_high-resolution_2022}, we hypothesize that our facial latent space does not retain substantial structural information of \( I \) after multiple layers of encoding in EMOCA \cite{danecek_emoca_2022}. Therefore, using 1D convolutions is sufficiently effective. 

The network is designed with convolutional ResNet blocks \cite{resnetblock}, each followed by a spatial attention block and then a temporal attention block. The noise \( \lambda_t \) is integrated into the ResNet block using scale and shift operators, while conditioning information \(c = (C , \mathcal{P} , \mathcal{E}) \) is encoded with a lightweight MLP encoder and incorporated into the spatial attention block via cross-attention. To embed temporal information, we scale and shift the features using temporal embeddings. Both the temporal and noise-time embeddings are encoded using sinusoidal embeddings for positional information. The up and down blocks have the architecture as illustrated in Fig. \ref{fig:unet-block}.

\textbf{Temporal Attention Layers.}
In general, this U-Net is similar to the standard spatial U-Net, with the key difference being the inclusion of temporal attention layers. Let \( z \in \mathbb{R}^{B \times T \times C \times D} \) represent the video latent vector, where \( D \) is the spatial latent dimension, \( C \) is the channel, \( T \) is time, and \( B \) is the batch size. The spatial layers treat each frame independently as a batch of size \( B \cdot T \), while the temporal layers operate on the temporal dimension, reinterpreting the latent vector as \( z \in \mathbb{R}^{B \times C \times T \times D} \) for processing. 

During training, we directly train the model using video data, rather than a mixture of images and videos as in \cite{ho_video_2022, blattmann_align_2023}, because the latent space is compact enough to allow the model to learn both spatial and temporal features simultaneously. It is not necessary to predict every frame, especially those that are very close to each other, as they share most features. To optimize inference time, we adopt the frame stacking approach from \cite{chen_diffusion_2024} that pushes close frames to the channel dimension to generate simultaneously.

\subsection{Elucidated Diffusion}
\label{ediff}


In this approach, we consider using the diffusion framework proposed by \cite{karras_elucidating_2022} as a more organized way to represent diffusion or score-based models, where the model is modeled as:

\begin{equation}
    D_\theta(\boldsymbol{z}, \sigma) = c_{\text {skip}}(\sigma) \boldsymbol{z} + c_{\text {out}}(\sigma) \cdot F_\theta\bigg(c_{\text {in}}(\sigma) \boldsymbol{z}, c_{\text {noise}}(\sigma), C \bigg),
\end{equation}

where \( \boldsymbol{z} \) is the facial latent vector, \( \sigma \) is the standard deviation of the Gaussian noise level, \( C \) represents conditions, and \( F_\theta \) is the temporal latent U-Net. By adjusting the terms \( c_{\text{skip}}, c_{\text{out}}, c_{\text{in}}, c_{\text{noise}} \), different diffusion strategies can be achieved with minimal changes to the model. We adopt the \( c \) terms from the Elucidated Diffusion Model (EDM) \cite{karras_elucidating_2022} due to the flexibility of the network architecture. As a result, we do not consider reparameterization approaches, since EDM \cite{karras_elucidating_2022} is a hybrid of both velocity, start, and noise prediction.

For clarity, with $y$ being the clear sample and $n$ being noise, the training objective is simplified as:

\begin{equation}
\mathbb{E}_{\boldsymbol{y} \sim p_{\text {data}},\boldsymbol{n} \sim \mathcal{N}\left(\mathbf{0}, \sigma^2 \mathbf{I}\right)}\bigg[\|\boldsymbol{y}_{\text{pred}} - \boldsymbol{y}_{\text{target}}\|_2^2\bigg],
\end{equation}

where

\begin{equation}
y_{\text{pred}} = F_\theta\bigg(c_{\text {in }}(\sigma) \cdot (\boldsymbol{y} + \boldsymbol{n}), c_{\text {noise }}(\sigma)\bigg),
\end{equation}

and

\begin{equation}
y_{\text{target}} = \frac{1}{c_{\text {out }}(\sigma)}\bigg(\boldsymbol{y} - c_{\text {skip }}(\sigma) \cdot (\boldsymbol{y} + \boldsymbol{n})\bigg).
\end{equation}

To guide the generation and combine different condition signals, we add the following guidance \cite{xu_vasa-1_2024, ho_classifier-free_2022} during inference:

\begin{equation}
\hat{z} = \left(1 + \sum_{c \in C} \lambda_{c}\right) \cdot F_\theta\left(z, t, C\right) - \sum_{c \in C} \lambda_{c} \cdot F_\theta\left(z, t,\left.C\right|_{c = \emptyset}\right),
\end{equation}

\begin{figure} \centering \includegraphics[width=0.4\linewidth]{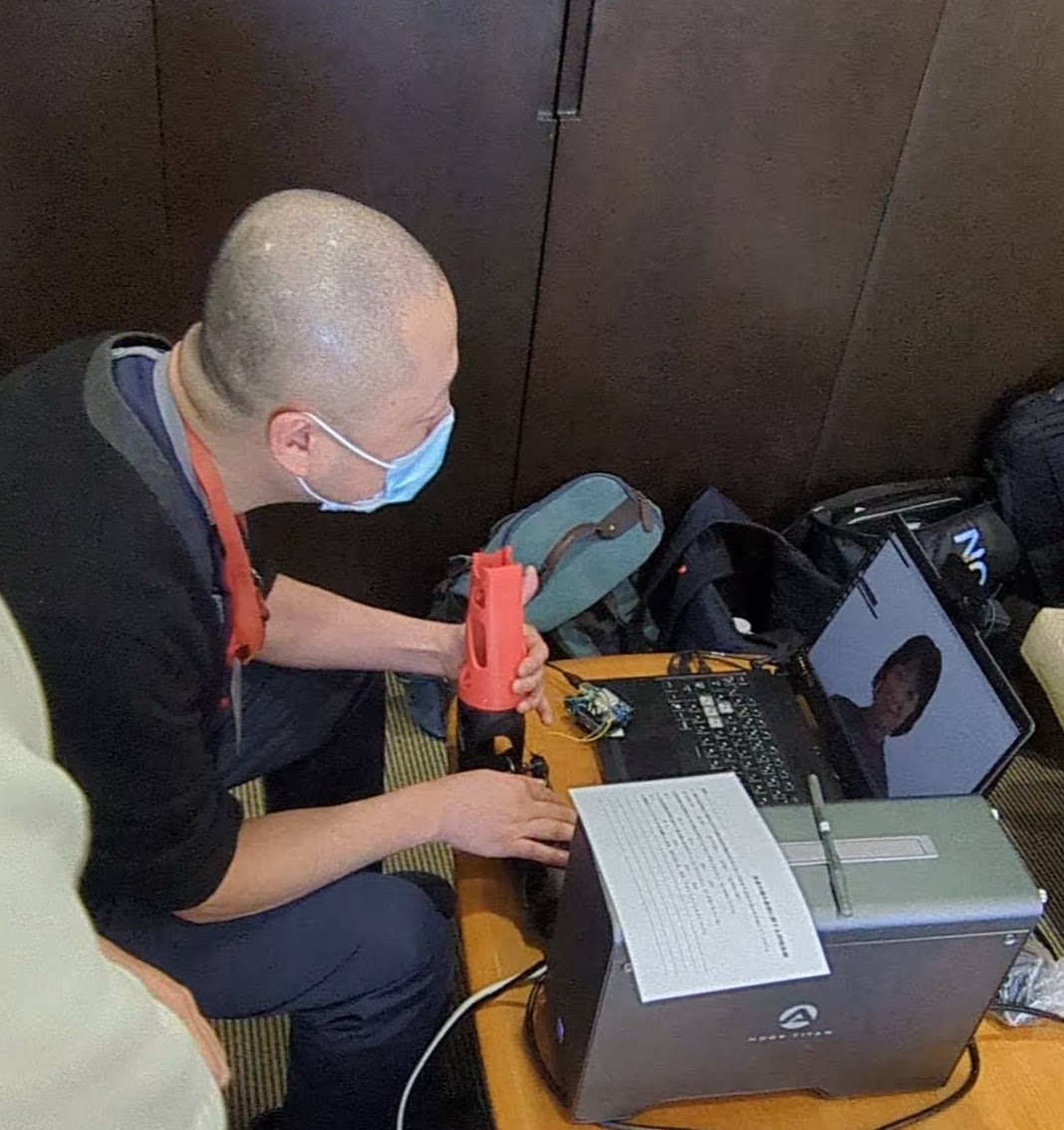} \caption{A clinician performing an elbow range-of-motion rehabilitation exercise while observing the virtual avatar's reaction.} \label{fig:clinician-robot} \end{figure}

where \(\lambda_c\) is the guidance strength of the condition. During training, we randomly drop each condition with a probability of 0.1 and search for optimal guidance strength for each condition. This denoising diffusion model works on a short sequence video \( Y \) because we consider physical pain facial expressions to be short-term behaviors, primarily focused on immediate stimuli signals and it speeds up the forward process. Therefore, we train the model on relatively short videos and conditions. We also randomly trim the conditions to account for the initial period when the condition is brief. During sampling, we use DPM++ \cite{lu2022dpm} to support speed-up guided sampling that enables sampling high-quality samples within from 15 to 20 denoising steps.







\subsection{Diffusion Forcing}
\label{dforcing}

The goal of this model is to be applied to a robot and generate arbitrary-length predictions. We adopt diffusion forcing \cite{chen_diffusion_2024} to extend beyond the short training horizon. Diffusion forcing assigns different noise levels to each temporal timestep, placing more uncertainty on future frames while reducing uncertainty for past frames. It also introduces a hyperparameter that controls uncertainty—higher values result in greater uncertainty, which in turn requires more denoising steps.


As diffusion forcing requires, we train our model with random noise levels for each temporal frame. During sampling, we apply a scheduling matrix that denoises a window of frames \( w \) and then shifts the window by a horizon step \( h \), ensuring there are some overlapping context frames \( w - h > 0 \). This overlap determines how quickly the model can respond to oncoming stimuli signals and how many past frames affect to the current generation. For example, with a small forward time, a sampling rate of 32\,Hz, and \( h = 16 \), the model would react with a delay of 0.5 seconds and it consider 16 frames as context frames.

\subsection{Elbow Range of Movement Exercise Robot}

We build a simple single-joint elbow robot using an MX-64 motor controlled by an Arduino. The joint’s angular range is linearly mapped to the heat stimulus values from the BioVid HeatPain Database, simulating a patient experiencing pain when flexing their elbow. To create a realistic facial representation, we employ Gaussian Avatars \cite{qian2024gaussianavatars} to map the 3D FLAME mesh \cite{li_learning_2017} onto a lifelike avatar. This setup is used to allow clinicians to assess the realism of the real-time rehabilitation exercise.








\section{EXPERIMENTS}

Our experiments are designed with two primary goals. First, we aim to demonstrate that our model outperforms the baseline and common approaches in facial expression generation. Second, we seek to show that our model can produce arbitrary-length predictions without divergence. To achieve these objectives, we propose a new set of metrics specifically for evaluating pain facial expressions and long-term prediction errors. We then conduct a human evaluation with both laypersons and clinicians to assess the model's naturalistic. 

\subsection{Dataset}






To create the pain facial expression dataset, we use the BioVid Heatpain Database part C \cite{walter_biovid_2013}, which captures heat pain responses from 87 subjects with 4 levels of pain intensity and 30 minutes of data, separated by pauses. The original dataset includes 3 modalities: frontal face video (25\,FPS), biomedical signals, and heat stimuli signals. We split the dataset into two subsets: 61 subjects for training and 26 subjects for validation. The validation subsets maintain an equal ratio of male and female participants, and include 5 low-expression and 21 normal-expression subjects. We filter the original validation videos to contain sequences that are close to the moment the pain stimuli signal changes its intensity. We synchronize all the modalities to have the same sampling rate with the video.

\textbf{Preprocessing.} To compute the latent representation for each video frame in the dataset, we use EMOCA \cite{danecek_emoca_2022} to extract the expression code \( \psi \) and jaw pose \( \theta_{\text{jaw}} \), while calculating the mean face for all other features. To determine the pain expression configuration for each subject, we calculate the Prkachin and Solomon Pain Intensity (PSPI) \cite{prkachin2008structure} by extracting action units (AUs) using the state-of-the-art GraphAU \cite{luo_learning_2022}. We then compute the average PSPI for each identity, referring to this as the pain expressiveness configuration. Additionally, we extract the emotion index for each subject using the HSEmotion emotion extractor \cite{hsemotion}, averaging across frames to create a consistent emotion configuration for each subject.

In its final form, the pain facial expression dataset consists of 4 modalities: facial expression parameters, pain stimuli signal, pain expressiveness configuration, and emotion configuration. Our model is trained to generate the facial expression parameters conditioned on the other modalities.

\subsection{Experiments setup}

\textbf{Quantitative metrics.} To compare the effectiveness of expressing pain, we draw from metrics used in multiple facial reaction generation \cite{song2023multiple} and human behavior generation \cite{barquero_belfusion_2023} to propose the following set of evaluation metrics on two modalities of PSPI and FLAME expression parameters:

\begin{itemize}
    \item Sim: Dynamic Time Warping (DTW) to measure the \textit{temporal signal similarity} between the generated sequence's PSPI signal and the ground truth PSPI signal under the same pain stimuli.
    \item Corr: Uses the Pearson Correlation Coefficient (PCC) to quantify the linear correlation between the generated PSPI signal and the ground truth PSPI signal.
    \item Dist: Uses Pairwise Mean Squared Error (MSE) to evaluate the difference between the generated expressions and the ground truth expressions.
    \item Divrs: MSE of multiple generated expressions under the same stimuli signals to assess the diversity of the generated outputs.
    \item Var: Measures the variance of generated expressions within the same sequence to evaluate how varied the expressions are in a single sequence.
\end{itemize}

\textbf{Baselines.} We establish three baselines to validate the model’s effectiveness:

\begin{itemize}
    \item Nearest Neighbor: Performs segment search in the training dataset to find the pain stimuli signal most similar to the current signal.
    \item Random Training Sequence: Returns a random sequence from the training dataset.
    \item Vector Quantized VAE and Autoregressive Model: We use the winning model from multiple appropriate facial reactions challenge - REACT Challenge 2024\footnote{\href{https://sites.google.com/cam.ac.uk/react2024/home}{The Second REACT Challenge@IEEE FG24}} \cite{dam2024finite} with modifications to take stimuli signal as input. For short, we refer to this method as autoregressive.
\end{itemize}

\begin{figure} \centering \includegraphics[width=0.7\linewidth]{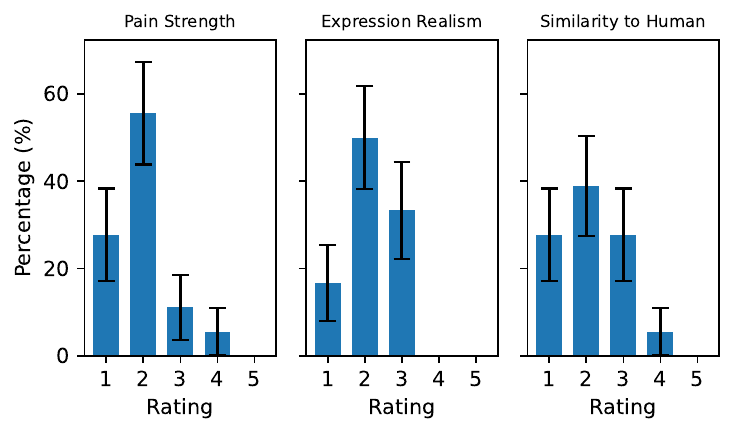} \caption{Clinicians' ratings of the virtual avatar’s quality after performing the elbow range-of-motion exercise. A rating of 1 indicates the lowest quality, while 5 indicates the highest.} \label{fig:arm-exercise-rating} \end{figure}

\textbf{Implementation details.} We train PainDiffusion with a sequence length of 64, a warm-up phase of 5k steps, and a total of 300k training steps, a learning rate of \(4 \times 10^{-4}\), and an exponential moving average (EMA) with a decay of 0.999. Training is conducted on a pair of NVIDIA 3080 GPUs. All the metrics and qualitative output are computed with a generation length of 640 frames, 10 times longer than the training horizon, to confirm the model's ability to generate arbitrary-length output. 

\textbf{Qualitative experiment setup.} To evaluate the realism of the generated pain expressions, we conduct a user study involving healthcare professionals. Specifically, we recruited 18 Japanese rehabilitation clinicians, all of whom regularly interact with patients experiencing pain. The experiment consists of two main phases: real-time interaction evaluation and video-based preference testing.

In the first phase, participants interacted with the robotic elbow for two minutes, observing the corresponding immediate facial reactions displayed on a screen. Figure \ref{fig:clinician-robot} illustrates the experimental setup for this phase. We ran a pilot experiment with 10 laypeople to decide which questions to ask. The clinicians were then asked to complete a questionnaire assessing the realism of the generated expressions across three key dimensions: (1) Response Dynamics: "How swift and strong is the reaction to the stimuli?"; (2) Motion Realism: "Do the generated movements appear realistic?"; (3) Patient Resemblance: "Does the reaction resemble that of a real patient?". Participant ratings for each question were recorded on a 5-point Likert scale.

The second phase involves a preference test using video-based comparisons. Participants were presented with 22 questions, assessing their preferences based on three aspects: (1) Temporal consistency with stimuli signals (2 questions), (2) Realism of facial reactions against groundtruth (8 questions), (3)
Diversity of facial expressions (9 questions). 

The experiment is implemented using the jsPsych framework\footnote{\href{https://www.jspsych.org/latest/}{jsPsych is a  behavioral experiments framework that run in a web browser.}}. From a dataset of 50 random validation video samples, each question is randomly selected from the dataset. To assess diversity, we generate four variations of the same stimuli. For both evaluating temporal consistency and diversity, participants rate the strength of the reaction on a 5-point scale. Each question is separated by a fixed screen in one second to inform the user of the boundary between the questions. We learn that cropping mouth and eye regions helps users better assess diversity as it limits the cognitive load of the comparison. We have 2/3 of diversity questions in the format of cropping eye and mouth.

\begin{figure}
    \centering
    \includegraphics[width=0.7\linewidth]{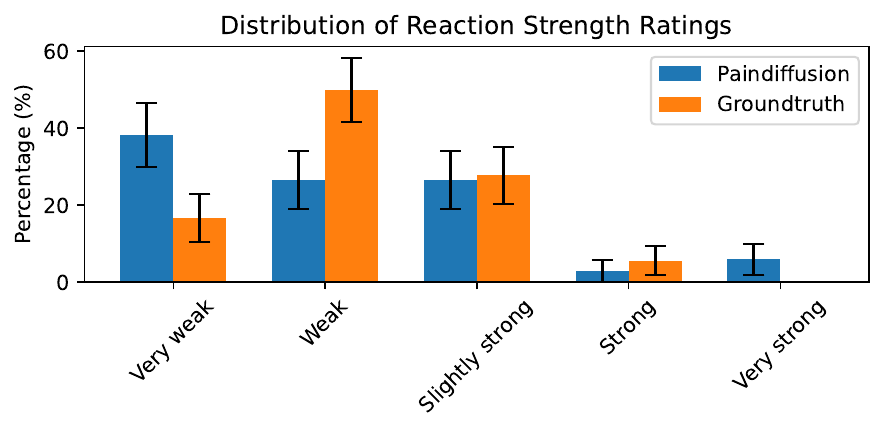}
    \caption{The distribution of clinicians' rating for temporal consistency with stimuli signal of PainDiffusion and Groundtruth in the video-preference experiment.}
    \label{fig:reaction-strength}
\end{figure}

\begin{figure*}
    \centering
    \includegraphics[width=1\linewidth]{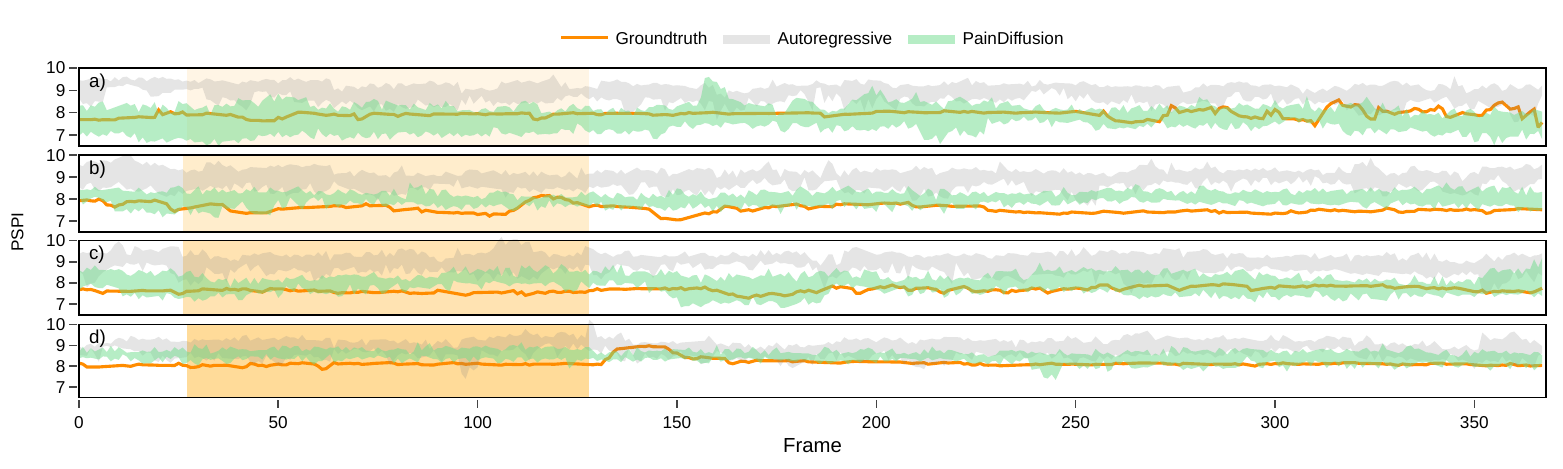}
    \caption{Visualization of four sample stimuli from the BioVid HeatPain Database validation set, ranging from level 1 (a) to level 4 (d) (darker orange indicates higher pain intensity). We ran five independent predictions for both PainDiffusion and the autoregressive baseline and plotted their range distributions as green and gray shaded areas, respectively. The orange-shaded region represents the duration during which the pain stimulus was applied to the subjects. PainDiffusion was configured with an emotion of contempt and an expression intensity of 8.5 on the PSPI scale \cite{prkachin2008structure}. The ground truth PSPI signal predominantly overlaps with the green-shaded region of PainDiffusion, whereas the autoregressive baseline tends to exhibit higher pain intensity on the PSPI scale.}
    \label{fig:distribution-graph}
\end{figure*}







\subsection{Qualitative results}











In the first phase of our qualitative evaluation, we analyzed the distribution of clinicians' ratings from real-time rehabilitation exercise experiments, as shown in Figure \ref{fig:arm-exercise-rating}. The results indicate that PainDiffusion generates relatively weak reactions, which aligns with the characteristics of the naturalistic dataset it was trained on. However, we acknowledge that clinicians expected reactions to align more closely with Japanese cultural norms, where expressions of pain tend to be more restrained. As anticipated, our model does not fully capture this cultural specificity, though it was judged to produce reactions that bear some resemblance to how Japanese patients typically express pain. Additionally, clinicians noted that the absence of a reference baseline for pain reactions made it difficult to assess whether the generated expressions were entirely appropriate. This highlights the need for calibration when deploying automated pain expression models in clinical applications to ensure cultural and contextual suitability.

\begin{figure}
    \centering
    \includegraphics[width=1\linewidth]{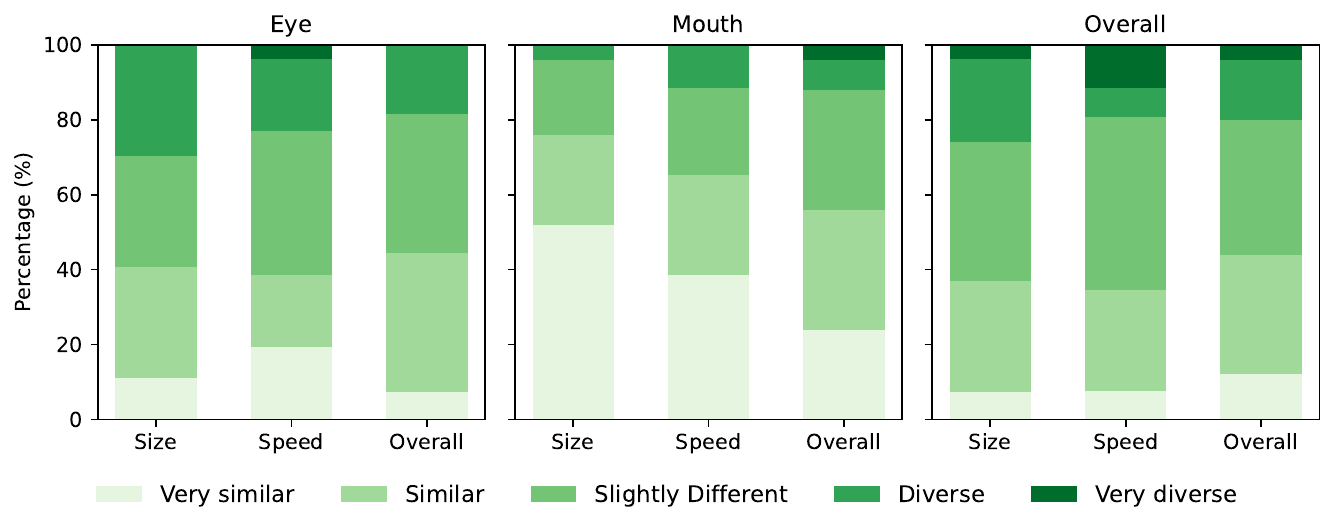}
    \caption{Diversity of generation of Paindiffusion assessed by clinicians.}
    \label{fig:diversity}
\end{figure}

In the second phase, the first video-based survey showed that PainDiffusion achieved a win rate of \textbf{31.2\% ± 4.8\%} against the ground truth in terms of perceived realism, suggesting that PainDiffusion is capable of expressing pain in a way that is convincing to human observers. Since naturalistic pain expression datasets typically exhibit weaker reactions \cite{moosaei_using_2017}, and the BioVid HeatPain Database reflects this characteristic, PainDiffusion remains consistent with observations in the dataset. However, as illustrated in Figure \ref{fig:reaction-strength}, the model exhibits slightly weaker temporal consistency compared to the ground truth. We further evaluated diversity along three dimensions: movement amplitude, movement type (fast or slow), and overall variability. As shown in Figure \ref{fig:diversity}, the model generates greater diversity in the eye region but lower diversity in the mouth region. Overall, it achieves moderate diversity across both movement size and type. These findings suggest that PainDiffusion is not only capable of generating realistic pain expressions but also has the potential to serve as a viable replacement for real patients in clinical training and simulation settings.

\begin{figure}
    \includegraphics[width=1.075\linewidth]{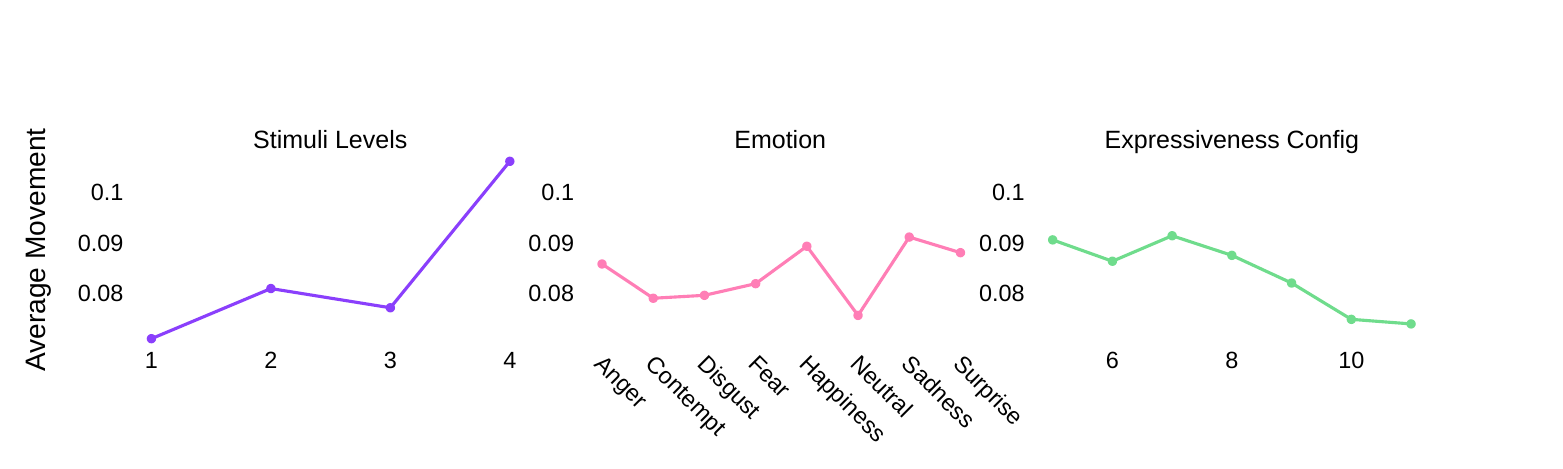}
    \caption{Average facial movement generated by PainDiffusion under varying stimuli levels, emotion configurations, and expressiveness configurations, while keeping other configurations constant. Higher stimuli levels correspond to greater movement, though different emotions exhibit varying levels of movement. Additionally, increased pain expressiveness tends to result in weaker overall movement.}
    \label{fig:avg-movement-different-config}
\end{figure}

We employed the Farneback method \cite{farneback2003two} to analyze facial motion patterns, quantifying both the magnitude and distribution of movements in the generated sequences. To evaluate controllability, we systematically varied three parameters: stimuli intensity, emotional state, and expressiveness settings. The resulting mean movements are visualized in Fig. \ref{fig:avg-movement-different-config}. As expected, increasing stimuli levels consistently produced more pronounced facial movements. Emotional states significantly modulated these responses - neutral states showed minimal movement, while sadness and happiness amplified pain-related facial motions. Notably, higher expressiveness settings (measured by average PSPI \cite{prkachin2008structure} scores) correlated with reduced movement intensity. This counterintuitive finding suggests that individuals who maintain higher baseline pain expressions in the dataset may actually exhibit more subtle changes in facial movement, possibly indicating a different strategy for communicating pain intensity.


\subsection{Quantitative results}

                 
                 

\begin{table*}[t]
\centering
\caption{\textbf{Baselines comparison.}}
\begin{tabular}{rll|lll}
\toprule
                 Modalities & \multicolumn{2}{c}{PSPI \cite{prkachin2008structure}} & \multicolumn{3}{c}{FLAME Params \cite{li_learning_2017}} \\ 
\cmidrule(l){2-6} 
                 Metrics & Sim  $\downarrow$  & Corr $\uparrow$ 
                 & Dist $\downarrow$ & Divrs $\uparrow$ & Var $\uparrow$ \\ 
                 & & ($10^{-3}$) &  & & \\
\midrule
Ground truth     & 0     & 999.9      & 0.00   & 0.00    & 0.06    \\
\midrule
\multicolumn{1}{r}{\textit{Naive Methods}} \\
Random Training Sample         & $353^{\scriptscriptstyle \pm 139.4}$   & $650^{\scriptscriptstyle \pm 5.0}$      & 0.28   & 0.23    & 0.02    \\
Nearest Neighbor & $342^{\scriptscriptstyle \pm 0}$   & $715^{\scriptscriptstyle \pm 0}$      & $0.27$   & $0.00$    & $0.03$    \\ 
\midrule
\multicolumn{1}{r}{\textit{Model-based Methods}} \\
FSQ-VAE Autoregressive \cite{dam2024finite} & $299^{\scriptscriptstyle \pm 1.17}$   & $396^{\scriptscriptstyle \pm 1.4}$    & 0.23    & 0.03     & 0.02    \\
PainDiffusion w/ Full-seq Diffusion & \textcolor{darkteal}{$218^{\scriptscriptstyle \pm 1.85}$}   & {$499^{\scriptscriptstyle \pm 2.6}$}    & $0.16$     & \textcolor{darkteal}{$0.09$}      & \textcolor{darkteal}{$0.05$}    \\
PainDiffusion w/ Diffusion Forcing   & $226^{\scriptscriptstyle \pm 0.89}$   & \textcolor{darkteal}{$597^{\scriptscriptstyle \pm 2.2}$}      & \textcolor{darkteal}{$0.10$}     & $0.06$      & $0.02$    \\  
\bottomrule 
\end{tabular}
\label{tab:resulttable}
\end{table*}

Table \ref{tab:resulttable} presents a comparison between our proposed method and other baseline approaches. Overall, our method outperforms the autoregressive baseline across all metrics. Notably, our evaluation methods provide a more detailed perspective on the pain generation problem by using metrics based on the PSPI signal from both the generated output and the ground truth. As illustrated in Fig. \ref{fig:distribution-graph}, the PSPI signal generated by PainDiffusion is more closely aligned with the ground truth compared to the autoregressive baseline, and heuristic baselines as evidenced by lower PainSim and higher PainCorr in both of its variants.

\begin{figure}
    \centering

    \begin{subfigure}[b]{0.29\textwidth}
        \centering
        \includegraphics[width=\linewidth]{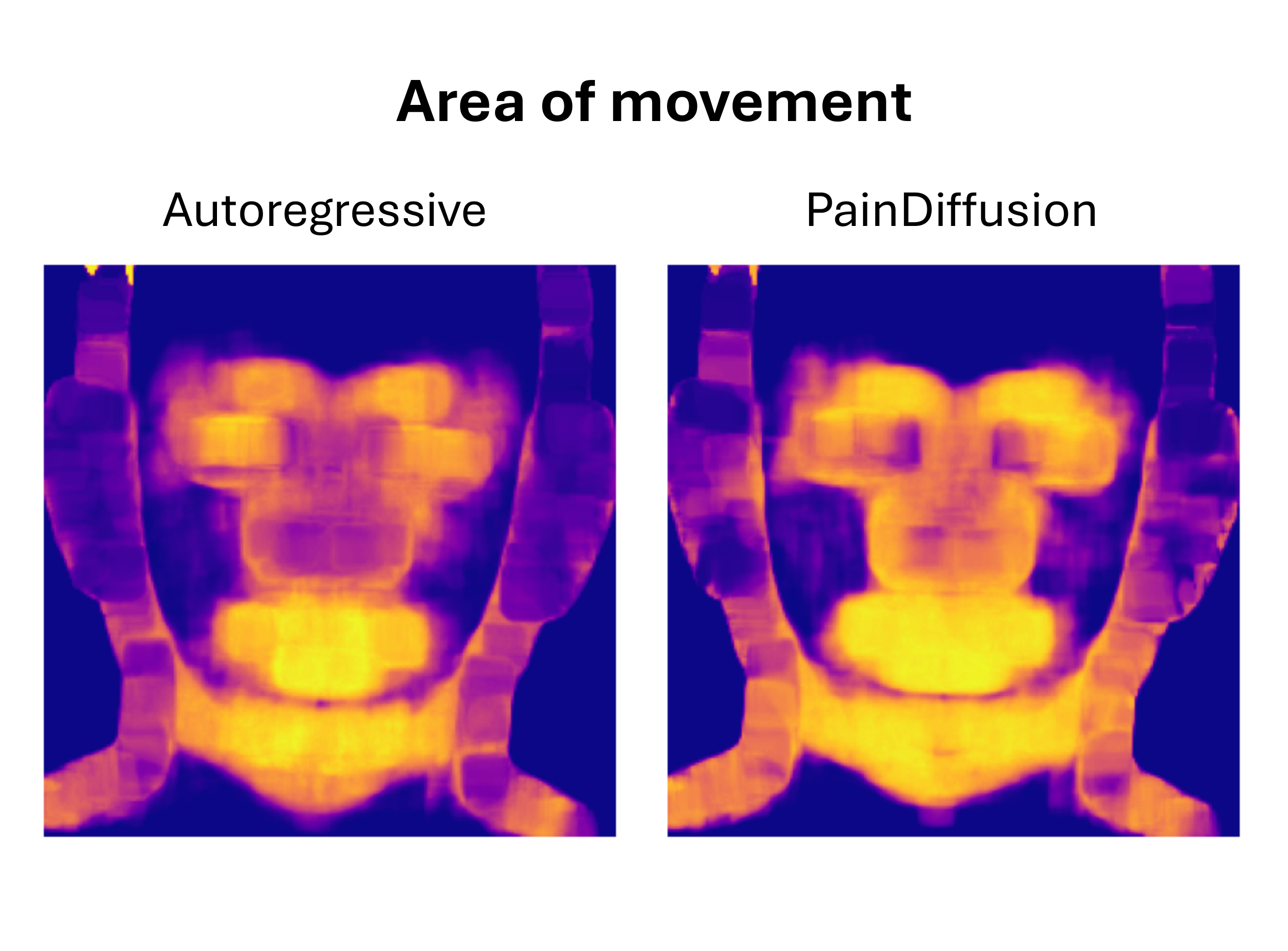}
    \end{subfigure}
    \caption{Average of area of movement of a validation sample from PainDiffusion and Autoregressive baseline.}
    \label{fig:qualitative-pd-vs-auto}
\end{figure}

For expression quality, PainDiffusion generates facial expressions that maintain better fidelity to ground truth while exhibiting greater diversity across multiple predictions, as indicated by lower Dist and higher Divrs metrics. Within individual sequences, our model achieves the highest expression variance (Var). Analysis using the Farneback method \cite{farneback2003two} reveals that PainDiffusion concentrates facial movements in key areas - mouth, eyebrows, chin, and nose - closely matching ground truth patterns (Fig. \ref{fig:qualitative-pd-vs-auto}).

In terms of real-time robotic applications, diffusion forcing proves particularly effective. While it shows slightly lower similarity scores and reduced diversity metrics (Divrs, Var) compared to full sequence diffusion, it achieves significantly better correlation and distance metrics. This makes diffusion forcing the more suitable choice for real-time robotic implementations where stable long-sequence generation is critical.


\subsection{Ablation}

We conducted ablation studies to explore the hyperparameter space of PainDiffusion. To optimize computational resources, we evaluated a subset of the validation set to compute the relevant metrics. The results are summarized in Table \ref{tab:ablationresult}. Our findings suggest that the context window size significantly impacts both temporal coherence and diversity in the generated expressions. From experiments with a window size of 64 frames, we observed that a context window of 16 frames achieves the best balance between temporal consistency and diversity. Additionally, a diffusion forcing uncertainty value of 2 yielded the most stable results. For guidance strengths, we found that setting values of (0.5, 1.00, 2.00) for emotion configuration, pain expression configuration, and stimulus signal, respectively, provides an optimal trade-off between PSPI metrics and expression diversity metrics.

\begin{table}[t]
\centering
\caption{\textbf{Ablation Study: Hyperparameters search.} }
\begin{tabular}{cccccccc}
\toprule
\multicolumn{3}{c}{Ablation}  & Sim    & Corr   & Dist & Divrs & Var \\ 

\multicolumn{3}{c}{}  &  & $10^{-3}$ & $10^{-1}$ & $10^{-1}$ & $10^{-1}$  \\
\midrule
\multicolumn{8}{l}{\textbf{Context Window Size}} \\
\multicolumn{3}{r}{8 frames}        & {305}   & \textcolor{darkteal}{639.1}     & 1.16   & \textcolor{darkteal}{0.68}    & \textcolor{darkteal}{0.26}   \\
\multicolumn{3}{r}{16 frames}        & \textcolor{darkteal}{295}   & 503.4   & 1.04   & 0.64    & 0.25  \\
\multicolumn{3}{r}{32 frames}        & 301   & {529.9}     & \textcolor{darkteal}{0.89}   & 0.59    & {0.25}  \\ \midrule
\multicolumn{8}{l}{\textbf{Diffusion Forcing Uncertainty}} \\
\multicolumn{3}{r}{0.5}      & {314}   & 422.0      & {1.57}   & \textcolor{darkteal}{0.84}    & \textcolor{darkteal}{0.44}   \\
\multicolumn{3}{r}{1}        & 307   & 505.5      & 1.40   & 0.74    & 0.35  \\
\multicolumn{3}{r}{2}        & \textcolor{darkteal}{279}   & \textcolor{darkteal}{624.8}      & {0.99}   & 0.63    & 0.25  \\
\multicolumn{3}{r}{4}        & 306   & 496.4      & \textcolor{darkteal}{0.95}   & 0.62    & 0.25  \\ \midrule
\multicolumn{8}{l}{\textbf{Guiding Strength}} \\
Emo. & Exp. & Sti. & & & & & \\
1.00 & 1.00 & 1.00         & 300   & \textcolor{darkteal}{592.7}      & \textcolor{darkteal}{0.85}   & 0.53    & 0.19   \\
1.00 & 2.00 & 4.00         & \textcolor{darkteal}{295}   & 353.1      & \textcolor{darkteal}{0.85}   & 0.54    & 0.19  \\
0.50 & 1.00 & 2.00         & {296}   & 558.6     & 0.97   & 0.61    & 0.22  \\
0.25 & 0.50 & 1.00         & 311   & {380.2}      & 1.02   & \textcolor{darkteal}{0.63}    & \textcolor{darkteal}{0.24}  \\ \bottomrule
\end{tabular}
\label{tab:ablationresult}
\end{table}

\section{CONCLUSION}

In this work, we introduced PainDiffusion, a novel model for generating appropriate and controllable facial expressions in response to pain stimuli. By leveraging diffusion forcing within a latent diffusion model, PainDiffusion effectively captures temporal dynamics to produce efficient, long-term predictions suitable for robotic applications. Our evaluation demonstrates that the model generates more diverse and specific expressions than baselines, achieving superior performance in pain-specific metrics like PSPI similarity and correlation.

Despite promising results, several areas for future work are crucial. The current model is limited to non-verbal expressions, with control restricted to emotional state and expressiveness. Furthermore, testing was only conducted on a virtual avatar, emphasizing the need for evaluation on a physical robot interface. Crucially, qualitative insights from clinicians highlight the necessity of culturally-aware, personalized medical technology. This finding points to a significant limitation of current datasets and the need for more diverse and naturalistic facial pain data.







\bibliographystyle{ieeetr}
\bibliography{root}

\begin{thebibliography}{10}

\bibitem{henry_association_2012}
S.~G. Henry, A.~Fuhrel-Forbis, M.~A.~M. Rogers, and S.~Eggly, ``Association between nonverbal communication during clinical interactions and outcomes: a systematic review and meta-analysis,'' {\em Patient Educ Couns}, vol.~86, pp.~297--315, Mar. 2012.

\bibitem{prkachin_expressing_1995}
K.~M. Prkachin and K.~D. Craig, ``Expressing pain: {The} communication and interpretation of facial pain signals,'' {\em Journal of Nonverbal Behavior}, vol.~19, no.~4, pp.~191--205, 1995.
\newblock Place: Germany Publisher: Springer.

\bibitem{giannini_measurement_2000}
A.~J. Giannini, J.~D. Giannini, and R.~K. Bowman, ``Measurement of nonverbal receptive abilities in medical students,'' {\em Percept Mot Skills}, vol.~90, pp.~1145--1150, June 2000.

\bibitem{jansen_emotional_2010}
J.~Jansen, J.~C.~M. van Weert, J.~de~Groot, S.~van Dulmen, T.~J. Heeren, and J.~M. Bensing, ``Emotional and informational patient cues: the impact of nurses' responses on recall,'' {\em Patient Educ Couns}, vol.~79, pp.~218--224, May 2010.

\bibitem{mcguire_comprehensive_1992}
D.~B. McGuire, ``Comprehensive and multidimensional assessment and measurement of pain,'' {\em J Pain Symptom Manage}, vol.~7, pp.~312--319, July 1992.

\bibitem{moosaei_using_2017}
M.~Moosaei, S.~K. Das, D.~O. Popa, and L.~D. Riek, ``Using {Facially} {Expressive} {Robots} to {Calibrate} {Clinical} {Pain} {Perception},'' in {\em Proceedings of the 2017 {ACM}/{IEEE} {International} {Conference} on {Human}-{Robot} {Interaction}}, (Vienna Austria), pp.~32--41, ACM, Mar. 2017.

\bibitem{pourebadi_facial_2022}
M.~Pourebadi and L.~D. Riek, ``Facial {Expression} {Modeling} and {Synthesis} for {Patient} {Simulator} {Systems}: {Past}, {Present}, and {Future},'' {\em ACM Trans. Comput. Healthcare}, vol.~3, pp.~23:1--23:32, Mar. 2022.

\bibitem{leonard_human_2004}
M.~Leonard, S.~Graham, and D.~Bonacum, ``The human factor: the critical importance of effective teamwork and communication in providing safe care,'' {\em Qual Saf Health Care}, vol.~13 Suppl 1, pp.~i85--90, Oct. 2004.

\bibitem{moosaei_naturalistic_2014}
M.~Moosaei, M.~J. Gonzales, and L.~D. Riek, ``Naturalistic {Pain} {Synthesis} for {Virtual} {Patients},'' in {\em Intelligent {Virtual} {Agents}} (T.~Bickmore, S.~Marsella, and C.~Sidner, eds.), (Cham), pp.~295--309, Springer International Publishing, 2014.

\bibitem{lee2021care}
M.~Lee, D.~T. Tran, H.~Yamazoe, and J.-H. Lee, ``Care training assistant robot and visual-based feedback for elderly care education environment,'' in {\em 2021 IEEE/SICE International Symposium on System Integration (SII)}, pp.~572--577, IEEE, 2021.

\bibitem{walter_biovid_2013}
S.~Walter, S.~Gruss, H.~Ehleiter, J.~Tan, H.~C. Traue, P.~Werner, A.~Al-Hamadi, S.~Crawcour, A.~O. Andrade, and G.~Moreira~da Silva, ``The biovid heat pain database data for the advancement and systematic validation of an automated pain recognition system,'' in {\em 2013 {IEEE} {International} {Conference} on {Cybernetics} ({CYBCO})}, pp.~128--131, June 2013.

\bibitem{haque_deep_2018}
M.~A. Haque, R.~B. Bautista, F.~Noroozi, K.~Kulkarni, C.~B. Laursen, R.~Irani, M.~Bellantonio, S.~Escalera, G.~Anbarjafari, K.~Nasrollahi, O.~K. Andersen, E.~G. Spaich, and T.~B. Moeslund, ``Deep {Multimodal} {Pain} {Recognition}: {A} {Database} and {Comparison} of {Spatio}-{Temporal} {Visual} {Modalities},'' in {\em 2018 13th {IEEE} {International} {Conference} on {Automatic} {Face} \& {Gesture} {Recognition} ({FG} 2018)}, pp.~250--257, May 2018.

\bibitem{huang2019pain}
D.~Huang, Z.~Xia, L.~Li, K.~Wang, and X.~Feng, ``Pain-awareness multistream convolutional neural network for pain estimation,'' {\em Journal of Electronic Imaging}, vol.~28, no.~4, pp.~043008--043008, 2019.

\bibitem{ng_learning_2022}
E.~Ng, H.~Joo, L.~Hu, H.~Li, T.~Darrell, A.~Kanazawa, and S.~Ginosar, ``Learning to listen: Modeling non-deterministic dyadic facial motion,'' in {\em Proceedings of the IEEE/CVF Conference on Computer Vision and Pattern Recognition}, pp.~20395--20405, 2022.

\bibitem{chen_humanmac_2023}
L.-H. Chen, J.~Zhang, Y.~Li, Y.~Pang, X.~Xia, and T.~Liu, ``{HumanMAC}: {Masked} {Motion} {Completion} for {Human} {Motion} {Prediction},'' in {\em 2023 {IEEE}/{CVF} {International} {Conference} on {Computer} {Vision} ({ICCV})}, (Paris, France), pp.~9510--9521, IEEE, Oct. 2023.

\bibitem{tian_transfusion_2024}
S.~Tian, M.~Zheng, and X.~Liang, ``{TransFusion}: {A} {Practical} and {Effective} {Transformer}-{Based} {Diffusion} {Model} for {3D} {Human} {Motion} {Prediction},'' {\em IEEE Robotics and Automation Letters}, vol.~9, pp.~6232--6239, July 2024.

\bibitem{barquero_belfusion_2023}
G.~Barquero, S.~Escalera, and C.~Palmero, ``Belfusion: Latent diffusion for behavior-driven human motion prediction,'' in {\em Proceedings of the IEEE/CVF International Conference on Computer Vision}, pp.~2317--2327, 2023.

\bibitem{kirschstein_diffusionavatars_2024}
T.~Kirschstein, S.~Giebenhain, and M.~Nie{\ss}ner, ``Diffusionavatars: Deferred diffusion for high-fidelity 3d head avatars,'' in {\em Proceedings of the IEEE/CVF Conference on Computer Vision and Pattern Recognition}, pp.~5481--5492, 2024.

\bibitem{chen_diffusion_2024}
B.~Chen, D.~Mart{\'\i}~Mons{\'o}, Y.~Du, M.~Simchowitz, R.~Tedrake, and V.~Sitzmann, ``Diffusion forcing: Next-token prediction meets full-sequence diffusion,'' 2025.

\bibitem{ho_classifier-free_2022}
J.~Ho and T.~Salimans, ``Classifier-free diffusion guidance,'' in {\em NeurIPS 2021 Workshop on Deep Generative Models and Downstream Applications}, 2021.

\bibitem{danecek_emoca_2022}
R.~Danecek, M.~Black, and T.~Bolkart, ``{EMOCA}: {Emotion} {Driven} {Monocular} {Face} {Capture} and {Animation},'' in {\em 2022 {IEEE}/{CVF} {Conference} on {Computer} {Vision} and {Pattern} {Recognition} ({CVPR})}, (New Orleans, LA, USA), pp.~20279--20290, IEEE, June 2022.

\bibitem{li_learning_2017}
T.~Li, T.~Bolkart, M.~J. Black, H.~Li, and J.~Romero, ``Learning a model of facial shape and expression from {4D} scans,'' {\em ACM Trans. Graph.}, vol.~36, pp.~1--17, Dec. 2017.

\bibitem{feng2021learning}
Y.~Feng, H.~Feng, M.~J. Black, and T.~Bolkart, ``Learning an animatable detailed 3d face model from in-the-wild images,'' {\em ACM Transactions on Graphics (ToG)}, vol.~40, no.~4, pp.~1--13, 2021.

\bibitem{pytorch3d}
J.~Johnson, N.~Ravi, J.~Reizenstein, D.~Novotny, S.~Tulsiani, C.~Lassner, and S.~Branson, ``Accelerating 3d deep learning with pytorch3d,'' in {\em SIGGRAPH Asia 2020 Courses}, SA '20, (New York, NY, USA), Association for Computing Machinery, 2020.

\bibitem{blattmann_align_2023}
A.~Blattmann, R.~Rombach, H.~Ling, T.~Dockhorn, S.~W. Kim, S.~Fidler, and K.~Kreis, ``Align your latents: High-resolution video synthesis with latent diffusion models,'' in {\em Proceedings of the IEEE/CVF Conference on Computer Vision and Pattern Recognition}, pp.~22563--22575, 2023.

\bibitem{xu_vasa-1_2024}
S.~Xu, G.~Chen, Y.-X. Guo, J.~Yang, C.~Li, Z.~Zang, Y.~Zhang, X.~Tong, and B.~Guo, ``{VASA}-1: {Lifelike} {Audio}-{Driven} {Talking} {Faces} {Generated} in {Real} {Time},'' Apr. 2024.
\newblock arXiv:2404.10667 [cs].

\bibitem{ho_video_2022}
J.~Ho, T.~Salimans, A.~Gritsenko, W.~Chan, M.~Norouzi, and D.~J. Fleet, ``Video diffusion models,'' {\em Advances in Neural Information Processing Systems}, vol.~35, pp.~8633--8646, 2022.

\bibitem{unet_2015}
O.~Ronneberger, P.~Fischer, and T.~Brox, ``U-net: Convolutional networks for biomedical image segmentation,'' {\em CoRR}, vol.~abs/1505.04597, 2015.

\bibitem{rombach_high-resolution_2022}
R.~Rombach, A.~Blattmann, D.~Lorenz, P.~Esser, and B.~Ommer, ``High-resolution image synthesis with latent diffusion models,'' in {\em Proceedings of the IEEE/CVF conference on computer vision and pattern recognition}, pp.~10684--10695, 2022.

\bibitem{resnetblock}
K.~He, X.~Zhang, S.~Ren, and J.~Sun, ``Deep residual learning for image recognition,'' in {\em Proceedings of the IEEE conference on computer vision and pattern recognition}, pp.~770--778, 2016.

\bibitem{karras_elucidating_2022}
T.~Karras, M.~Aittala, T.~Aila, and S.~Laine, ``Elucidating the design space of diffusion-based generative models,'' in {\em Advances in neural information processing systems}, vol.~35, pp.~26565--26577, 2022.

\bibitem{lu2022dpm}
C.~Lu, Y.~Zhou, F.~Bao, J.~Chen, C.~Li, and J.~Zhu, ``Dpm-solver++: Fast solver for guided sampling of diffusion probabilistic models,'' {\em arXiv preprint arXiv:2211.01095}, 2022.

\bibitem{qian2024gaussianavatars}
S.~Qian, T.~Kirschstein, L.~Schoneveld, D.~Davoli, S.~Giebenhain, and M.~Nie{\ss}ner, ``Gaussianavatars: Photorealistic head avatars with rigged 3d gaussians,'' in {\em Proceedings of the IEEE/CVF Conference on Computer Vision and Pattern Recognition}, pp.~20299--20309, 2024.

\bibitem{prkachin2008structure}
K.~M. Prkachin and P.~E. Solomon, ``The structure, reliability and validity of pain expression: Evidence from patients with shoulder pain,'' {\em Pain}, vol.~139, no.~2, pp.~267--274, 2008.

\bibitem{luo_learning_2022}
C.~Luo, S.~Song, W.~Xie, L.~Shen, and H.~Gunes, ``Learning {Multi}-dimensional {Edge} {Feature}-based {AU} {Relation} {Graph} for {Facial} {Action} {Unit} {Recognition},'' in {\em Proceedings of the {Thirty}-{First} {International} {Joint} {Conference} on {Artificial} {Intelligence}}, pp.~1239--1246, July 2022.
\newblock arXiv:2205.01782 [cs].

\bibitem{hsemotion}
A.~V. Savchenko, L.~V. Savchenko, and I.~Makarov, ``Classifying emotions and engagement in online learning based on a single facial expression recognition neural network,'' {\em IEEE Transactions on Affective Computing}, vol.~13, no.~4, pp.~2132--2143, 2022.

\bibitem{song2023multiple}
S.~Song, M.~Spitale, Y.~Luo, B.~Bal, and H.~Gunes, ``Multiple appropriate facial reaction generation in dyadic interaction settings: What, why and how?,'' {\em arXiv preprint arXiv:2302.06514}, 2023.

\bibitem{dam2024finite}
Q.~T. Dam, T.~T.~N. Nguyen, D.~T. Tran, and J.-H. Lee, ``Finite scalar quantization as facial tokenizer for dyadic reaction generation,'' in {\em 2024 IEEE 18th International Conference on Automatic Face and Gesture Recognition (FG)}, pp.~1--5, IEEE, 2024.

\bibitem{farneback2003two}
G.~Farneb{\"a}ck, ``Two-frame motion estimation based on polynomial expansion,'' in {\em Image Analysis: 13th Scandinavian Conference, SCIA 2003 Halmstad, Sweden, June 29--July 2, 2003 Proceedings 13}, pp.~363--370, Springer, 2003.

\end{thebibliography}

\end{document}